\documentclass[10pt,twocolumn,letterpaper]{article}

\usepackage{cvpr}
\usepackage{times}
\usepackage{epsfig}
\usepackage{graphicx}
\usepackage{amsmath}
\usepackage{amssymb}
\usepackage{ulem}
\usepackage[linesnumbered,algoruled,boxed,lined]{algorithm2e}
\usepackage[numbers, sort, comma, square]{natbib}
\usepackage{subcaption}
\usepackage{authblk}

\cvprfinalcopy 

\def\cvprPaperID{6340} 

\ifcvprfinal\fi

\begin{document}

\title{Edge-Labeling Graph Neural Network for Few-shot Learning}


\author[1,3]{Jongmin Kim\thanks{Work done during an internship at Kakao Brain. Correspondence to  \texttt{kimjm0309@gmail.com}}}
\author[2,3]{Taesup Kim}
\author[3]{Sungwoong Kim}
\author[1]{Chang D.Yoo}

\affil[1]{Korea Advanced Institute of Science and Technology} 
\affil[2]{MILA, Universit\'e de Montr\'eal} 
\affil[3]{Kakao Brain}

\maketitle
\thispagestyle{empty}

\begin{abstract}
In this paper, we propose a novel edge-labeling graph neural network (EGNN), which adapts a deep neural network on the edge-labeling graph, for few-shot learning. The previous graph neural network (GNN) approaches in few-shot learning have been based on the node-labeling framework, which implicitly models the intra-cluster similarity and the inter-cluster dissimilarity. In contrast, the proposed EGNN learns to predict the edge-labels rather than the node-labels on the graph that enables the evolution of an explicit clustering by iteratively updating the edge-labels with direct exploitation of both intra-cluster similarity and the inter-cluster dissimilarity. It is also well suited for performing on various numbers of classes without retraining, and can be easily extended to perform a transductive inference. The parameters of the EGNN are learned by episodic training with an edge-labeling loss to obtain a well-generalizable model for unseen low-data problem. On both of the supervised and semi-supervised few-shot image classification tasks with two benchmark datasets, the proposed EGNN significantly improves the performances over the existing GNNs.
\end{abstract}

\section{Introduction}
A lot of interest in meta-learning \cite{Lemke15} has been recently arisen in various areas including especially task-generalization problems such as few-shot learning \cite{vinyals2016matching, snell2017prototypical, finn2017model, yang2018learning, garcia2017few, ren2018meta, ravi2016optimization, Santoro2016mann, MishraICLR18, Borisnips18, Yanbin18, Wang18, Lake15, kim2018bayesian}, learn-to-learn \cite{AndrychowiczDGH16, Bello17, Wichrowska17}, non-stationary reinforcement learning\cite{MaruanICLR18, Houthoofnips18, Clavera18}, and continual learning \cite{Vuorio18, Ju}. Among these meta-learning problems, few-shot leaning aims to automatically and efficiently solve new tasks with few labeled data based on knowledge obtained from previous experiences. This is in contrast to traditional (deep) learning methods that highly rely on large amounts of labeled data and cumbersome manual tuning to solve a single task.

Recently, there has also been growing interest in graph neural networks (GNNs) to handle rich relational structures on data with deep neural networks \cite{Battaglia18, Bronstein, xu2018powerful, GilmerSRVD17, Gori05, Scarselli09, Kipf17, Li16, Hamilton17, Velickovic18, Defferrard16}. GNNs iteratively perform a feature aggregation from neighbors by message passing, and therefore can express complex interactions among data instances. Since few-shot learning algorithms have shown to require full exploitation of the relationships between a support set and a query \cite{MishraICLR18, Borisnips18, snell2017prototypical, vinyals2016matching, yang2018learning}, the use of GNNs can naturally have the great potential to solve the few-shot learning problem.
\begin{figure}[t]
\centering
\includegraphics[width=0.47\textwidth]{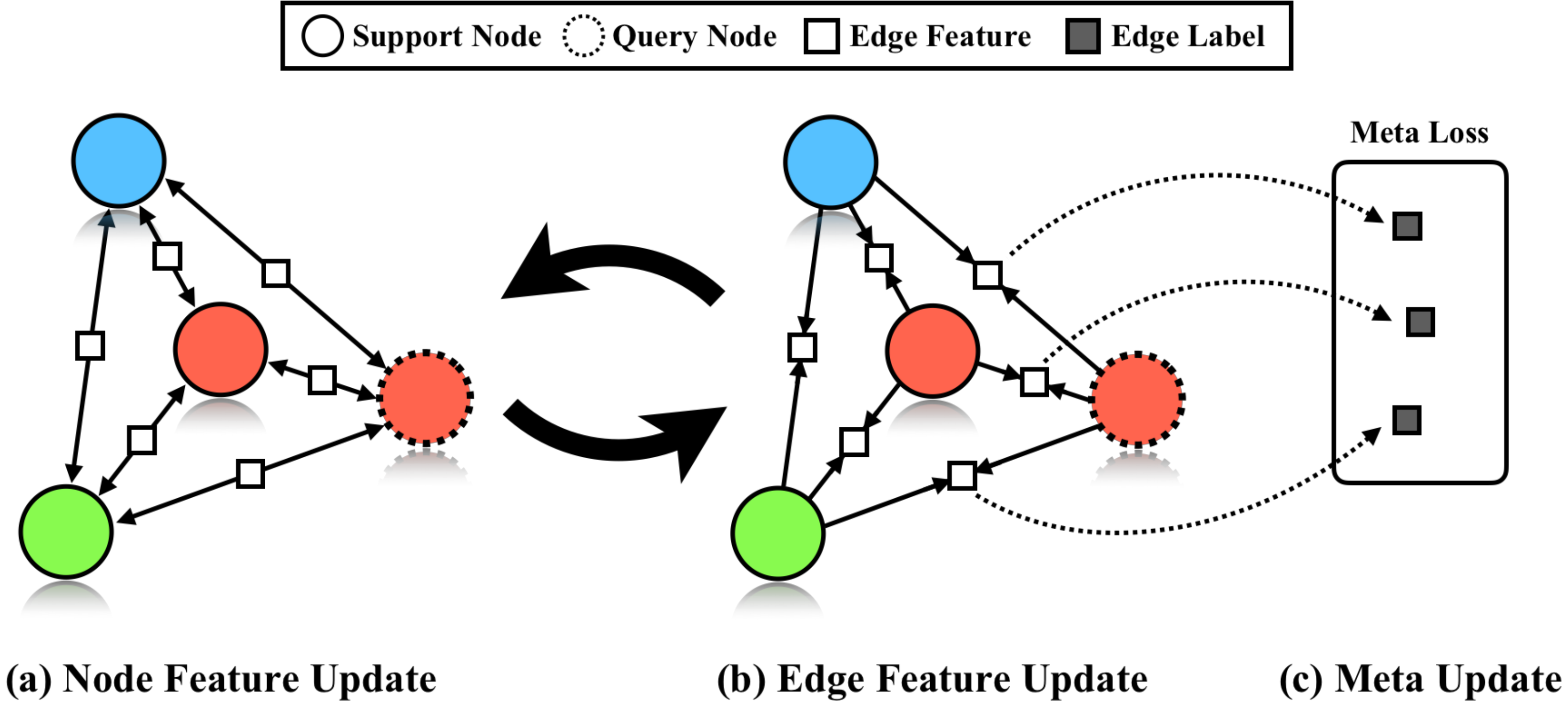}
\caption{Alternative node and edge feature update in EGNN with edge-labeling for few-shot learning}
\label{fig:EGNN_update}
\end{figure}
A few approaches that have explored GNNs for few-shot learning have been recently proposed \cite{garcia2017few, Yanbin18}. Specifically, given a new task with its few-shot support set, \citet{garcia2017few} proposed to first construct a graph where all examples of the support set and a query are densely connected. Each input node is represented by the embedding feature (e.g. an output of a convolutional neural network) and the given label information (e.g. one-hot encoded label). 
Then, it classifies the unlabeled query by iteratively updating node features from neighborhood aggregation. \citet{Yanbin18} proposed a transductive propagation network (TPN) on the node features obtained from a deep neural network. At test-time, it iteratively propagates one-hot encoded labels over the entire support and query instances as a whole with a common graph parameter set. Here, it is noted that the above previous GNN approaches in few-shot learning have been mainly based on the node-labeling framework, which {\it implicitly} models the intra-cluster similarity and inter-cluster dissimilarity. 

On the contrary, the edge-labeling framework is able to {\it explicitly} perform the clustering with representation learning and metric learning, and thus it is intuitively a more conducive framework for inferring a query association to an existing support clusters. Furthermore, it does not require the pre-specified number of clusters (e.g. class-cardinality or ways) while the node-labeling framework has to separately train the models according to each number of clusters. The explicit utilization of edge-labeling which indicates whether the associated two nodes belong to the same cluster (class) have been previously adapted in the naive (hyper) graphs for correlation clustering \cite{kim2011higher} and the GNNs for citation networks or dynamical systems \cite{gong2018adaptive, kipf2018neural}, but never applied to a graph for few-shot learning. Therefore, in this paper, we propose an edge-labeling GNN (EGNN) for few-shot leaning, especially on the task of few-shot classification.

The proposed EGNN consists of a number of layers in which each layer is composed of a node-update block and an edge-update block. Specifically, across layers, the EGNN not only updates the node features but also explicitly adjusts the edge features, which reflect the edge-labels of the two connected node pairs and directly exploit both the intra-cluster similarity and inter-cluster dissimilarity. As shown in Figure \ref{fig:EGNN_update}, after a number of alternative node and edge feature updates, the edge-label prediction can be obtained from the final edge feature. The edge loss is then computed to update the parameters of EGNN with a well-known meta-learning strategy, called episodic training \cite{Santoro2016mann, vinyals2016matching}. The EGNN is naturally able to perform a transductive inference to predict all test (query) samples at once as a whole, and this has shown more robust predictions in most cases when a few labeled training samples are provided. In addition, the edge-labeling framework in the EGNN enables to handle various numbers of classes without remodeling or retraining. We will show by means of experimental results on two benchmark few-shot image classification datasets that the EGNN outperforms other few-shot learning algorithms including the existing GNNs in both supervised and semi-supervised cases.

Our main contributions can be summarized as follows:
\begin{itemize}
\item The EGNN is first proposed for few-shot learning with iteratively updating edge-labels with exploitation of both intra-cluster similarity and inter-cluster dissimilarity. It is also able to be well suited for performing on various numbers of classes without retraining.
\item It consists of a number of layers in which each layer is composed of a node-update block and an edge-update block where the corresponding parameters are estimated under the episodic training framework.
\item Both of the transductive and non-transductive learning or inference are investigated with the proposed EGNN.
\item On both of the supervised and semi-supervised few-shot image classification tasks with two benchmark datasets, the proposed EGNN significantly improves the performances over the existing GNNs. Additionally, several ablation experiments show the benefits from the explicit clustering as well as the separate utilization of intra-cluster similarity and inter-cluster dissimilarity.
\end{itemize}



\section{Related works}
\label{sec:related works}
\paragraph{Graph Neural Network} Graph neural networks were first proposed to directly process graph structured data with neural networks as of form of recurrent neural networks \cite{Gori05, Scarselli09}. \citet{Li16} further extended it with gated recurrent units and modern optimization techniques. Graph neural networks mainly do representation learning with a neighborhood aggregation framework that the node features are computed by recursively aggregating and transforming features of neighboring nodes. Generalized convolution based propagation rules also have been directly applied to graphs \cite{Bruna13, Henaff15, Defferrard16}, and \citet{Kipf17} especially applied it to semi-supervised learning on graph-structured data with scalability. A few approaches \cite{garcia2017few, Yanbin18} have explored GNNs for few-shot learning and are based on the node-labeling framework.
\paragraph{Edge-Labeling Graph} Correlation clustering (CC) is a graph-partitioning algorithm \cite{Bansal04} that infers the edge labels of the graph by simultaneously maximizing intra-cluster similarity and inter-cluster dissimilarity. \citet{Finley05} considered a framework that uses structured support vector machine in CC for noun-phrase clustering and news article clustering. \citet{Taskar04} derived a max-margin formulation for learning the edge scores in CC for producing two different segmentations of a single image. \citet{kim2011higher} explored a higher-order CC over a hypergraph for task-specific image segmentation. The attention mechanism in a graph attention network has recently extended to incorporate real-valued edge features that are adaptive to both the local contents and the global layers for modeling citation networks \cite{gong2018adaptive}. \citet{kipf2018neural} introduced a method to simultaneously infer relational structure with interpretable edge types while learning the dynamical model of an interacting system.
\citet{johnson2016learning} introduced the Gated Graph Transformer Neural Network (GGT-NN) for natural language tasks, where multiple edge types and several graph transformation operations including node state update, propagation and edge update are considered.

\paragraph{Few-Shot Learning} One main stream approach for few-shot image classification is based on representation learning and does prediction by using nearest-neighbor according to similarity between representations. The similarity can be a simple distance function such as cosine or Euclidean distance. A Siamese network~\cite{Koch15} works in a pairwise manner using trainable weighted $L_{1}$ distance.  A matching network~\cite{vinyals2016matching} further uses an attention mechanism to derive an differentiable nearest-neighbor classifier and a prototypical network~\cite{snell2017prototypical} extends it with defining prototypes as the mean of embedded support examples for each class. DEML \cite{DEML} has introduced a concept learner to extract high-level concept by using a large-scale auxiliary labeled dataset showing that a good representation is an important component to improve the performance of few-shot image classification.

A meta-learner that learns to optimize model parameters extract some transferable knowledge between tasks to leverage in the context of few-shot learning. Meta-LSTM~\cite{ravi2016optimization} uses LSTM as a model updater and treats the model parameters as its hidden states. This allows to learn the initial values of parameters and update the parameters by reading few-shot examples. MAML~\cite{finn2017model} learns only the initial values of parameters and simply uses SGD. It is a model agnostic approach, applicable to both supervised and reinforcement learning tasks. Reptile~\cite{reptile} is similar to MAML but using only first-order gradients.  Another generic meta-learner, SNAIL~\cite{MishraICLR18}, is with a novel combination of temporal convolutions and soft attention to learn an optimal learning strategy.

\section{Method}
\label{sec:method}
In this section, the definition of few-shot classification task is  introduced, and the proposed algorithm is described in detail. 
\subsection{Problem definition: Few-shot classification}
The few-shot classification aims to learn a classifier when only a few training samples per each class are given. Therefore, each few-shot classification {\it task} ${\mathcal T}$ contains a {\it support set ${\mathcal S}$}, a labeled set of input-label pairs, and a {\it query set ${\mathcal Q}$}, an unlabeled set on which the learned classifier is evaluated. If the support set ${\mathcal S}$ contains $K$ labeled samples for each of $N$ unique classes, the problem is called {\it $N$-way $K$-shot} classification problem. 

Recently, meta-learning has become a standard methodology to tackle few-shot classification. In principle, we can train a classifier to assign a class label to each query sample with only the compact support set of the task. However, a small number of labeled support samples for each task are not sufficient to train a model fully reflecting the inter- and intra-class variations, which often leads to unsatisfactory classification performance. Meta-learning on explicit training set resolves this issue by extracting transferable knowledge that allows us to perform better few-shot learning on the support set, and thus classify the query set more successfully.

As an efficient way of meta-learning, we adopt {\it episodic training} \cite{Santoro2016mann, vinyals2016matching} which is commonly employed in various literatures \cite{snell2017prototypical, finn2017model, yang2018learning}. Given a relatively large labeled training dataset, the idea of episodic training is to sample {\it training tasks (episodes)} that mimic the few-shot learning setting of test tasks. Here, since the distribution of training tasks is assumed to be similar to that of test tasks, the performances of the test tasks can be improved by learning a model to work well on the training tasks.

More concretely, in episodic training, both training and test tasks of the $N$-way $K$-shot problem are formed as follows: ${\mathcal T} = {\mathcal S}  \bigcup {\mathcal Q}$ where ${\mathcal S} = \{({\bf x}_i, y_i)\}_{i=1}^{N \times K}$ and ${\mathcal Q} =\{({\bf x}_i, y_i)\}_{i=N \times K + 1}^{N \times K + T}$. Here, $T$ is the number of query samples, and ${\bf x}_i$ and $y_i \in \{C_1, \cdots C_N\} = {\mathcal C}_{\mathcal T} \subset {\mathcal C}$ are the $i$th input data and its label, respectively. ${\mathcal C}$ is the set of all classes of either training or test dataset. Although both the training and test tasks are sampled from the common task distribution, the label spaces are mutually exclusive, i.e. ${\mathcal C}_{train} \cap {\mathcal C}_{test} = \emptyset$. The support set ${\mathcal S}$ in each episode serves as the labeled training set on which the model is 
trained to minimize the loss of its predictions over the query set ${\mathcal Q}$. This training procedure is iteratively carried out episode by episode until convergence.

Finally, if some of $N \times K$ support samples are unlabeled, the problem is referred to as {\it semi-supervised} few-shot classification. In Section \ref{sec: Experiment}, the effectiveness of our algorithm on semi-supervised setting will be presented. 

\subsection{Model} 
\label{subsec: model}
\begin{figure*}
\begin{center}
\includegraphics[width=1.0\linewidth]{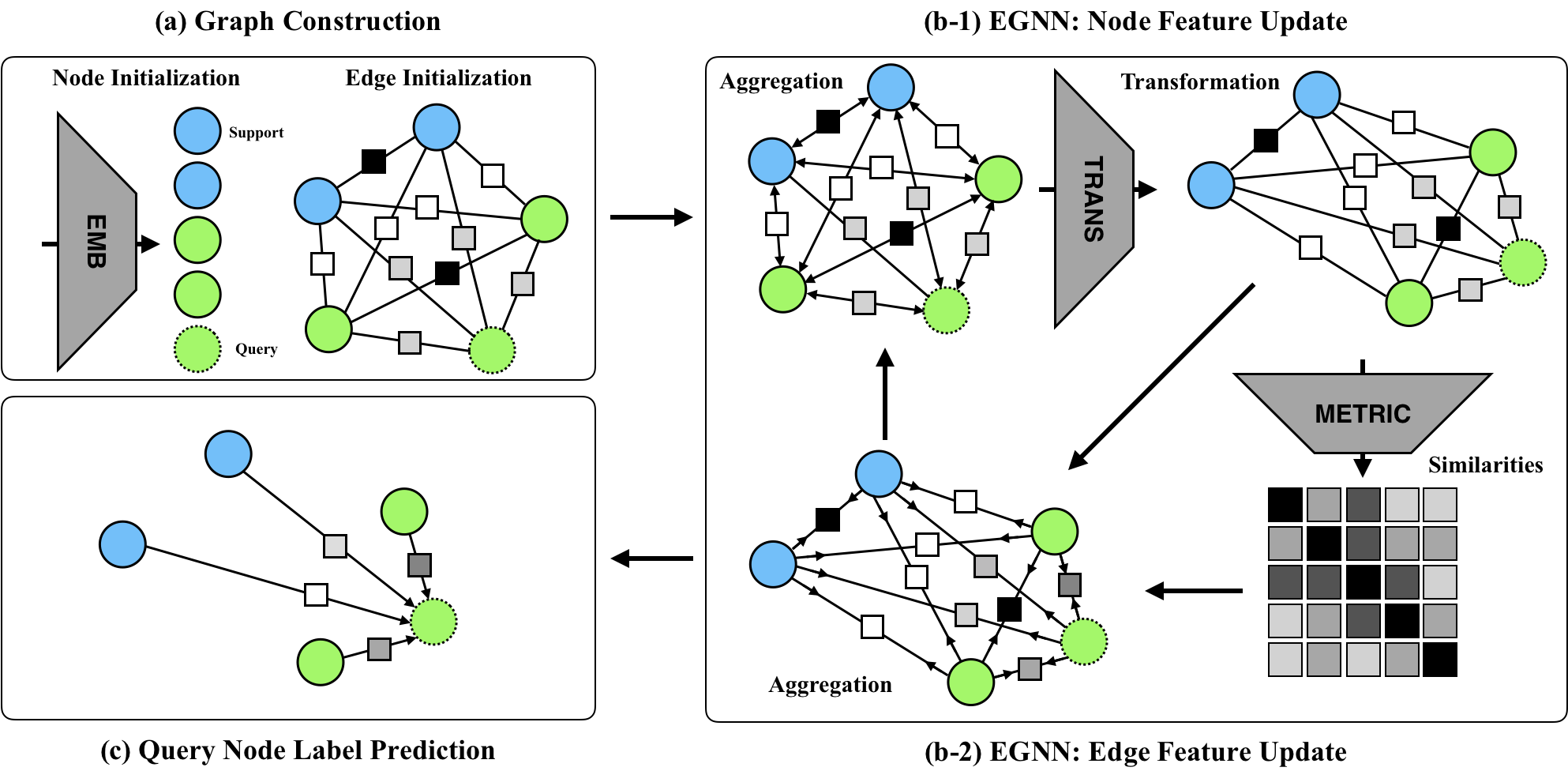}
\end{center}
\vspace{-0.5cm}
   \caption{The overall framework of the proposed EGNN model. In this illustration, a 2-way 2-shot problem is presented as an example. Blue and green circles represent two different classes. Nodes with solid line represent labeled support samples, while a node with dashed line represents the unlabeled query sample. The strength of edge feature is represented by the color in the square. Note that although each edge has a 2-dimensional feature, only the first dimension is depicted for simplicity. The detailed process is described in Section \ref{subsec: model}.}
\vspace{-0.5cm}
\label{fig:EGNN_overall}
\end{figure*}

This section describes the proposed EGNN for few-shot classification, as illustrated in Figure \ref{fig:EGNN_overall}. Given the feature representations (extracted from a jointly trained convolutional neural network) of all samples of the target task, a fully-connected graph is initially constructed where each node represents each sample, and each edge represents the types of relationship between the two connected nodes; Let ${\mathcal G} = ({\mathcal V},{\mathcal E}; {\mathcal T})$ be the graph constructed with samples from the task $\mathcal T$, where ${\mathcal V}:=\{V_i\}_{i=1,...,|{\mathcal T}|}$ and ${\mathcal E}:=\{E_{ij}\}_{i,j = 1,...,|{\mathcal T}|}$ denote the set of nodes and edges of the graph, respectively. Let ${\bf v}_i$ and ${\bf e}_{ij}$ be the node feature of $V_i$ and the edge feature of $E_{ij}$, respectively. $|{\mathcal T}| = N \times K + T$ is the total number of samples in the task $\mathcal T$. Each ground-truth edge-label $y_{ij}$ is defined by the ground-truth node labels as:
\begin{equation}
y_{ij} = \left\{\begin{array}{cc} 1,& \mbox{if} ~~ y_i = y_j, \\ 0,& \mbox{otherwise}.
\end{array}\right.
\end{equation}

Each edge feature ${\bf e}_{ij} = \{e_{ijd}\}_{d=1}^{2} \in [0, 1]^2$ is a 2-dimensional vector representing the (normalized) strengths of the intra- and inter-class relations of the two connected nodes. This allows to separately exploit the intra-cluster similarity and the inter-cluster dissimilairity. 

\begin{figure}[t]
\begin{center}
\hspace*{-0.5cm}
\includegraphics[width=1.1\linewidth]{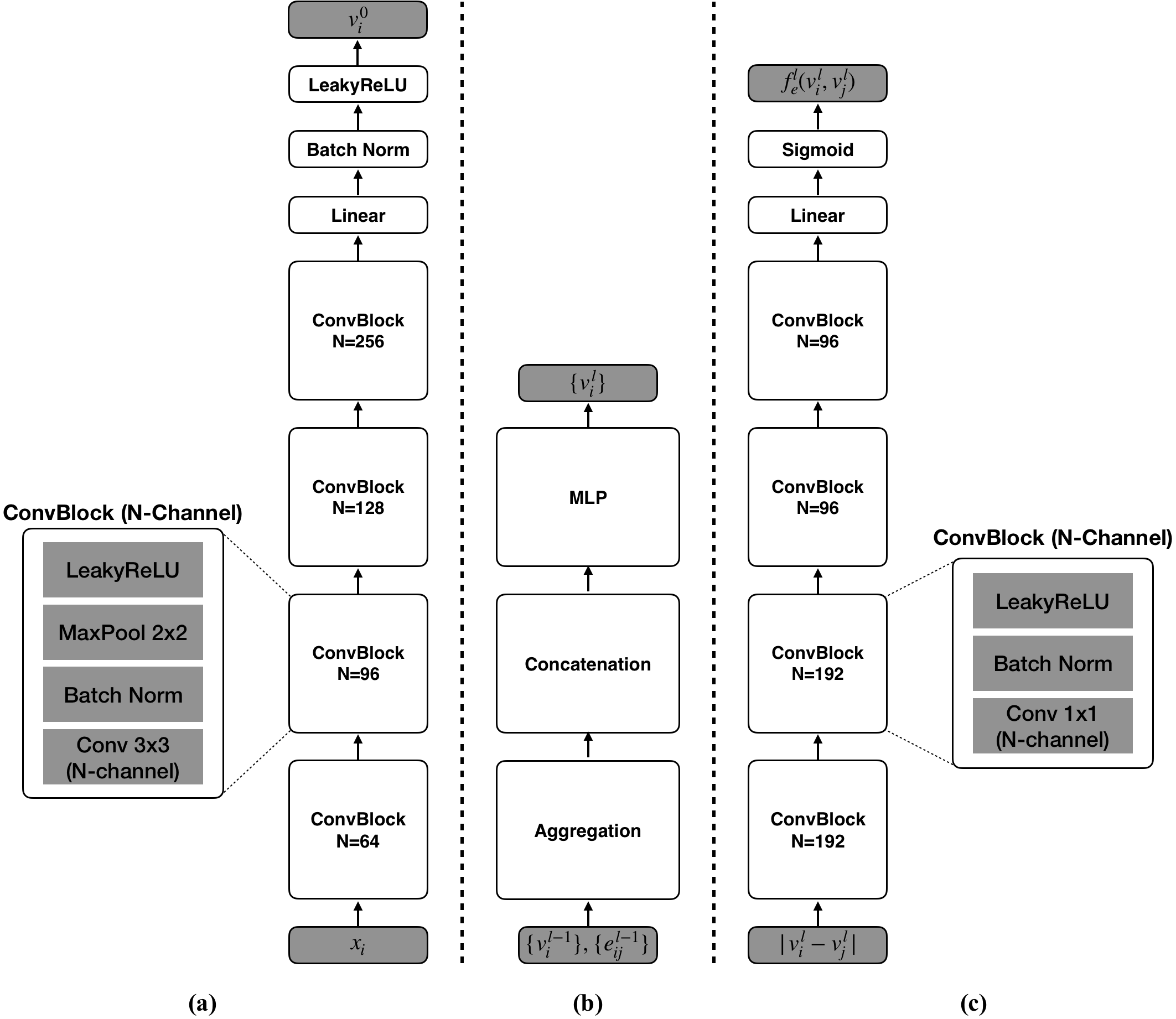}
\end{center}
   \caption{Detailed network architectures used in EGNN. (a) Embedding network $f_{emb}$. (b) Feature (node) transformation network $f_v^{\ell}$. (c) Metric network $f_e^{\ell}$.}
\vspace*{-0.5cm}   
\label{fig:network}
\end{figure}

Node features are initialized by the output of the convolutional embedding network ${\bf v}_i^0 = f_{emb}({\bf x}_i; \theta_{emb})$, where $\theta_{emb}$ is the corresponding parameter set (see Figure \ref{fig:network}.(a)). Edge features are initialized by edge labels as follows:
\begin{equation}
\hspace{-0.1cm}
{\bf e}_{ij}^0 = \left\{\begin{array}{cc} {[1||0]},& \mbox{if} ~ y_{ij} = 1 ~~ \mbox{and}~~ i, j \le N \times K, \\ {[0||1]},&  \mbox{if} ~ y_{ij} = 0 ~~ \mbox{and}~~ i, j \le N \times K, \\ {[0.5||0.5]},& \mbox{otherwise}, \end{array} \right.
\label{eq: edge init}
\end{equation}
where $||$ is the concatenation operation.

The EGNN consists of $L$ layers to process the graph, and the forward propagation of EGNN for inference is an alternative update of node feature and edge feature through layers. 

In detail, given ${\bf v}_i^{\ell-1}$ and ${\bf e}_{ij}^{\ell-1}$ from the layer $\ell-1$, \textit{node feature update} is firstly conducted by a neighborhood aggregation procedure. The feature node ${\bf v}_i^{\ell}$ at the layer $\ell$ is updated by first aggregating the features of other nodes proportional to their edge features, and then performing the feature transformation; the edge feature ${\bf e}_{ij}^{\ell-1}$ at the layer $\ell-1$ is used as a degree of contribution of the corresponding neighbor node like an attention mechanism as follows:
\begin{equation}
{\bf v}_{i}^{\ell} = f_v^{\ell}([\sum_{j} {\tilde e}_{ij1}^{\ell-1}{\bf v}_j^{\ell-1}||\sum_{j} {\tilde e}_{ij2}^{\ell-1}{\bf v}_j^{\ell-1}];\theta_v^{\ell}),
\label{eq: node update}
\end{equation}
where ${\tilde e}_{ijd} = \frac{e_{ijd}}{\sum_{k}e_{ikd}}$, and $f_v^{\ell}$ is the feature (node) transformation network, as shown in Figure \ref{fig:network}.(b), with the parameter set $\theta_v^{\ell}$. It should be noted that besides the conventional intra-class aggregation, we additionally consider inter-class aggregation. While the intra-class aggregation provides the target node the information of ``similar neighbors'', the inter-class aggregation provides the information of ``dissimilar neighbors''.

Then, \textit{edge feature update} is done based on the newly updated node features. The (dis)similarities between every pair of nodes are re-obtained, and the feature of each edge is updated by combining the previous edge feature value and the updated (dis)similarities such that
\begin{eqnarray}
{\bar e}_{ij1}^{\ell} &=& \frac{f_{e}^{\ell}({\bf v}_{i}^{\ell}, {\bf v}_{j}^{\ell}; \theta_e^{\ell})e_{ij1}^{\ell-1}}{\sum_{k}f_{e}^{\ell}({\bf v}_{i}^{\ell}, {\bf v}_{k}^{\ell}; \theta_e^{\ell})e_{ik1}^{\ell-1}/(\sum_{k}e_{ik1}^{\ell-1})}, \\ 
{\bar e}_{ij2}^{\ell} &=& \frac{(1-f_{e}^{\ell}({\bf v}_{i}^{\ell}, {\bf v}_{j}^{\ell}; \theta_e^{\ell}))e_{ij2}^{\ell-1}}{\sum_{k}(1-f_{e}^{\ell}({\bf v}_{i}^{\ell}, {\bf v}_{k}^{\ell}; \theta_e^{\ell}))e_{ik2}^{\ell-1}/(\sum_{k}e_{ik2}^{\ell-1})},
\\
{\bf e}_{ij}^{\ell} &=& {\bar {\bf e}}_{ij}^{\ell} / \|{\bar {\bf e}}_{ij}^{\ell}\|_1,
\label{eq:edge feature update}
\end{eqnarray}
where $f_{e}^{\ell}$ is the metric network that computes similarity scores with the parameter set $\theta_e^{\ell}$ (see Figure \ref{fig:network}.(c)). In specific, the node feature flows into edges, and each element of the edge feature vector is updated separately from each normalized intra-cluster similarity or inter-cluster dissimilarity. Namely, each edge update considers not only the relation of the corresponding pair of nodes but also the relations of the other pairs of nodes. We can optionally use two separate metric networks for the computations of each of similarity or dissimilarity (e.g. separate $f_{e,dsim}$ instead of $(1-f_{e,sim})$).

After $L$ number of alternative node and edge feature updates, the edge-label prediction can be obtained from the final edge feature, i.e. ${\hat y}_{ij} = e_{ij1}^L$. Here, ${\hat y}_{ij} \in [0,1]$ can be considered as a probability that the two nodes $V_i$ and $V_j$ are from the same class. Therefore, each node $V_i$ can be classified by simple weighted voting with support set labels and edge-label prediction results. The prediction probability of node $V_i$ can be formulated as $P(y_i = {\mathcal C}_k|{\mathcal T}) = p_i^{(k)}$:
\begin{equation}
 p_i^{(k)} =  \mathrm{softmax} \Big( \sum_{\{j:j \neq i \wedge ({\bf x}_j, y_j) \in {\mathcal S} \}} \!\!{\hat y}_{ij} \delta(y_j = {\mathcal C}_k) \Big)
\label{eq: node clssification}
\end{equation}
where $\delta(y_j = {\mathcal C}_k)$ is the Kronecker delta function that is equal to one when $y_j = {\mathcal C}_k$ and zero otherwise. 
Alternative approach for node classification is the use of graph clustering; the entire graph ${\mathcal G}$ can be first partitioned into clusters, using the edge prediction and an optimization for valid partitioning via linear programming \cite{kim2011higher}, and then each cluster can be labeled with the support label it contains the most. However, in this paper, we simply apply Eq. (\ref{eq: node clssification}) to obtain the classification results.
The overall algorithm for the EGNN inference at test-time is summarized in Algorithm 1. The non-transductive inference means the number of query samples $T$ = 1 or it performs the query inference one-by-one, separately, while the transductive inference classifies all query samples at once in a single graph.

\begin{algorithm}[t]
\caption{The process of EGNN for inference}
{\bf Input:}  ${\mathcal G} = ({\mathcal V},{\mathcal E}; {\mathcal T})$, where ${\mathcal T} = {\mathcal S}  \bigcup {\mathcal Q}$, ${\mathcal S} = \{({\bf x}_i, y_i)\}_{i=1}^{N \times K}$, ${\mathcal Q} =\{{\bf x}_i\}_{i=N \times K + 1}^{N \times K + T}$ \

{\bf Parameters:} $\theta_{emb} \cup \{\theta_v^{\ell}, \theta_{e}^{\ell}\}_{\ell=1}^{L}$\

{\bf Output:} $\{{\hat y}_{i}\}_{i=N \times K + 1}^{N \times K + T}$\

{\bf Initialize:}  ${\bf v}_i^0 = f_{emb}({\bf x}_i; \theta_{emb})$, ${\bf e}_{ij}^0, ~\forall i,j $\

\For{$\ell = 1,\cdots,L$}{
    \tcc{Node feature update}
    \For{$i = 1,\cdots, |V|$}{
    ${\bf v}_i^{\ell} \leftarrow {\tt NodeUpdate} (\{{\bf v}_i^{\ell-1}\}, \{{\bf e}_{ij}^{\ell-1}\}; \theta_v^{\ell})$\
    }
    \tcc{Edge feature update}
    \For{$(i,j) = 1,\cdots, |E|$}{
    ${\bf e}_{ij}^{\ell} \leftarrow {\tt EdgeUpdate} (\{{\bf v}_i^{\ell}\}, \{{\bf e}_{ij}^{\ell-1}\}; \theta_e^{\ell})$\
    }
}
\tcc{Query node label prediction}
$\{{\hat y}_{i}\}_{i=N \times K + 1}^{N \times K + T} \leftarrow {\tt Edge2NodePred}(\{y_{i}\}_{i=1}^{N \times K}, \{{\bf e}_{ij}^L\})$
\end{algorithm}

\subsection{Training}
Given $M$ training tasks $\{{\mathcal T}_{m}^{train}\}_{m=1}^{M}$ at a certain iteration during the episodic training, the parameters of the proposed EGNN, $\theta_{emb} \cup \{\theta_v^{\ell}, \theta_{e}^{\ell}\}_{\ell=1}^{L}$, are trained in an end-to-end fashion by minimizing the following loss function:
\begin{eqnarray}
{\mathcal L} = \sum_{{\ell}=1}^L \sum_{m=1}^{M} \lambda_{\ell} {\mathcal L}_{e}(Y_{m,e}, {\hat Y}_{m,e}^{\ell}), \label{eq: loss}
\end{eqnarray}
where $Y_{m,e}$ and ${\hat Y}_{m,e}^{\ell}$ are the set of all ground-truth query edge-labels and the set of all (real-valued) query-edge predictions of the $m^{\text{th}}$ task at the $\ell^{\text{th}}$ layer, respectively, and the edge loss ${\mathcal L}_e$ is defined as binary cross-entropy loss. 
Since the edge prediction results can be obtained not only from the last layer but also from the other layers, the total loss combines all losses that are computed in all layers in order to improve the gradient flow in the lower layers. 

\section{Experiments} \label{sec: Experiment}
We evaluated and compared our EGNN \footnote{The code and models are available on \tt{https://github.com/khy0809/fewshot-egnn}.} 
with state-of-the-art approaches on two few-shot learning benchmarks, i.e. {\it mini}ImageNet \cite{vinyals2016matching} and {\it tiered}ImageNet \cite{ren2018meta}. 
\subsection{Datasets}
\paragraph{\textbf {\textit {mini}ImageNet}}
 It is the most popular few-shot learning benchmark proposed by \cite{vinyals2016matching} derived from the original ILSVRC-12 dataset \cite{russakovsky2015imagenet}. All images are RGB colored, and of  size 84 $\times$ 84 pixels, sampled from 100 different classes with 600 samples per class. We followed the splits used in \cite{ravi2016optimization} - 64, 16, and 20 classes for training, validation and testing, respectively.
\paragraph{\textbf {\textit {tiered}ImageNet}} Similar to \textit{mini}ImageNet dataset, \textit{tiered}ImageNet \cite{ren2018meta} is also a subset of ILSVRC-12 \cite{russakovsky2015imagenet}. Compared with \textit{mini}ImageNet, it has much larger number of images (more than 700K) sampled from larger number of classes (608 classes rather than 100 for \textit{mini}ImageNet). Importantly, different from \textit{mini}ImageNet, \textit{tiered}ImageNet adopts hierarchical category structure where each of 608 classes belongs to one of 34 higher-level categories sampled from the high-level nodes in the Imagenet. Each higher-level category contains 10 to 20 classes, and divided into 20 training (351 classes), 6 validation (97 classes) and 8 test (160 classes) categories. The average number of images in each class is 1281.

\subsection{Experimental setup}
\paragraph{\bf Network Architecture} For feature embedding module, a convolutional neural network, which consists of four blocks, was utilized as in most few-shot learning models \cite{vinyals2016matching, finn2017model, snell2017prototypical, garcia2017few} without any skip connections \footnote{Resnet-based models are excluded for fair comparison.}. More concretely, each convolutional block consists of 3 $\times$ 3 convolutions, a batch normalization and a LeakyReLU activation. All network architectures used in EGNN are described in details in Figure \ref{fig:network}.
\paragraph{\bf Evaluation} For both datasets, we conducted a 5-way 5-shot experiment which is one of standard few-shot learning settings. For evaluation, each test episode was formed by randomly sampling 15 queries for each of 5 classes, and the performance is averaged over 600 randomly generated episodes from the test set. Especially, we additionally conducted a more challenging 10-way experiment on \textit{mini}Imagenet, to demonstrate the flexibility of our EGNN model when the number of classes are different between meta-training stage and meta-test stage, which will be presented in Section \ref{subsec: ablation_studies}.

\paragraph{\bf Training} The proposed model was trained with Adam optimizer with an initial learning rate of $5 \times 10^{-4}$ and weight decay of $10^{-6}$. The task mini-batch sizes for meta-training were set to be 40 and 20 for 5-way and 10-way experiments, respectively. For \textit{mini}ImageNet, we cut the learning rate in half every 15,000 episodes while for tieredImageNet, the learning rate is halved for every 30,000 because it is larger dataset and requires more iterations to converge.  All our code was implemented in Pytorch \cite{paszke2017automatic} and run with NVIDIA Tesla P40 GPUs.

\subsection{Few-shot classification}
The few-shot classification performance of the proposed EGNN model is compared with several state-of-the-art models in Table \ref{table: result_mini} and \ref{table: result_tiered}. Here, as presented in \cite{Yanbin18}, all models are grouped into three categories with regard to three different transductive settings; ``No'' means non-transductive method, where each query sample is predicted independently from other queries, ``Yes'' means transductive method where all queries are simultaneously processed and predicted together, and ``BN'' means that query batch statistics are used instead of global batch normalization parameters, which can be considered as a kind of transductive inference at test-time.

\begin{table}[t] 
\begin{subtable}{\linewidth}
\centering
\caption{\textit{mini}ImageNet}
\begin{tabular}{lcc}
\hline
Model                         & Trans. & 5-Way 5-Shot         \\ \hline
Matching Networks \cite{vinyals2016matching}            & No           & 55.30                             \\
Reptile \cite{reptile}                       & No           & 62.74                           \\
Prototypical Net \cite{snell2017prototypical}             & No           & 65.77                         \\
GNN \cite{garcia2017few}                     & No           & 66.41                            \\
{\bf EGNN}                 & No           & {\bf 66.85}                           \\
\hline
MAML    \cite{finn2017model}   & BN          & 63.11                     \\
Reptile + BN  \cite{reptile}                  & BN           & 65.99                          \\
Relation Net \cite{yang2018learning} & BN           & 67.07                    \\
\hline
MAML+Transduction \cite{finn2017model}            & Yes          & 66.19                        \\
TPN \cite{Yanbin18}                          & Yes          & 69.43                         \\
TPN (Higher $K$)  \cite{Yanbin18}           & Yes          & 69.86                        \\ 

{\bf EGNN+Transduction}    & Yes          & {\bf 76.37}                           \\ \hline
\end{tabular}
\label{table: result_mini}
\end{subtable}

\vspace{0.2cm}
\begin{subtable}{\linewidth}
\centering
\caption{\textit{tiered}ImageNet}
\begin{tabular}{lcc}
\hline
Model                         & Trans. & 5-Way 5-Shot         \\ \hline
Reptile \cite{reptile}                      & No           &      66.47                       \\
Prototypical Net    \cite{snell2017prototypical}          & No           &      69.57                      \\
{\bf EGNN}           & No           &      {\bf 70.98}                        \\
\hline
MAML \cite{finn2017model}                     & BN           &      70.30                   \\
Reptile + BN    \cite{reptile}              & BN           &      71.03                       \\
Relation Net \cite{yang2018learning}                  & BN           &     71.31                       \\ \hline
MAML+Transduction       \cite{finn2017model}      & Yes          &      70.83                       \\
TPN \cite{Yanbin18}                          & Yes          &      72.58                       \\

{\bf EGNN+Transduction}    & Yes          &   {\bf 80.15}                          \\ \hline
\end{tabular}
\label{table: result_tiered}
\end{subtable}
\caption{Few-shot classification accuracies on \textit{mini}ImageNet and \textit{tiered}ImageNet. All results are averaged over 600 test episodes.
Top results are highlighted.}
\vspace{-0.5cm}
\end{table}

The proposed EGNN was tested with both transductive and non-transductive settings. As shown in Table \ref{table: result_mini}, EGNN shows the best performance in 5-way 5-shot setting, on both transductive and non-transductive settings on \textit{mini}Imagenet. Notably, EGNN performed better than node-labeling GNN \cite{garcia2017few}, which supports the effectiveness of our edge-labeling framework for few-shot learning. Moreover, EGNN with transduction (EGNN + Transduction) outperformed the second best method (TPN \cite{Yanbin18}) on both datasets, especially by large margin on \textit{mini}Imagenet. Table \ref{table: result_tiered} shows that the transductive setting on \textit{tiered}Imagenet gave the best performance as well as large improvement compared to the non-transductive setting. In TPN, only the labels of the support set are propagated to the queries based on the pairwise node feature affinities using a common Laplacian matrix, so the queries communicate to each other only via their embedding feature similarities. In contrast, our proposed EGNN allows us to consider more complicated interactions between query samples, by propagating to each other not only their node features but also edge-label information across the graph layers having different parameter sets. Furthermore, the node features of TPN are fixed and never changed during label propagation, which allows them to derive a closed-form, one-step label propagation equation. On the contrary, in our EGNN, both node and edge features are dynamically changed and adapted to the given task gradually with several update steps. 

\subsection{Semi-supervised few-shot classification}
For semi-supervised experiment, we followed the same setting described in \cite{garcia2017few} for fair comparison. It is a 5-way 5-shot setting, but the support samples are only partially labeled. The labeled samples are balanced among classes so that all classes have the same amount of labeled and unlabeled samples.
The obtained results on \textit{mini}Imagenet are presented in Table \ref{table:semi_supervised}. Here, {\it ``LabeledOnly''} denotes learning with only labeled support samples, and {\it ``Semi''} means the semi-supervised setting explained above. Different results are presented according to when 20\% and 40\%, 60\% of support samples were labeled, and the proposed EGNN is compared with node-labeling GNN \cite{garcia2017few}. As shown in Table \ref{table:semi_supervised}, semi-supervised learning increases the performances in comparison to labeled-only learning on all cases. Notably, the EGNN outperformed the previous GNN \cite{garcia2017few} by a large margin (61.88\% vs 52.45\%, when 20\% labeled) on semi-supervised learning, especially when the labeled portion was small. The performance is even more increased on transductive setting (EGNN-Semi(T)). In a nutshell, our EGNN is able to extract more useful information from unlabeled samples compared to node-labeling framework, on both transductive and non-transductive settings. 

\begin{table}[]
\begin{center}
\begin{tabular}{l|cccc}
\hline
               & \multicolumn{4}{c}{Labeled Ratio (5-way 5-shot)} \\
Training method         & 20\%     & 40\%    &60\%  &   100\%    \\ \hline
GNN-LabeledOnly \cite{garcia2017few}  & 50.33         &    56.91      &    -       &66.41     \\
GNN-Semi    \cite{garcia2017few}   & 52.45          &   58.76       & -      &66.41        \\
EGNN-LabeledOnly & 52.86          &   -       & -      &66.85        \\
{\bf EGNN-Semi}      & {\bf 61.88}          &  {\bf 62.52}        & {\bf 63.53}           &66.85      \\ \hline
EGNN-LabeledOnly(T) &   59.18        &    -      &  -        &76.37        \\
{\bf EGNN-Semi(T)}      & {\bf 63.62}         & {\bf 64.32}         & {\bf 66.37}     &   76.37     \\ \hline
\end{tabular}
\caption{Semi-supervised few-shot classification accuracies on \textit{mini}ImageNet.}
\label{table:semi_supervised}
\vspace{-0.2cm}
\end{center}
\end{table}

\begin{table}[]
\begin{center}
\begin{tabular}{l|ccc}
\hline
& \multicolumn{3}{c}{\# of EGNN layers} \\
Feature type          & 1     & 2    & 3     \\ \hline
{\bf Intra \& Inter}  &  67.99         &  73.19        &  {\bf 76.37}             \\
Intra Only       & 67.28          & 72.20         & 74.04             \\
\hline
\end{tabular}
\caption{5-way 5-shot results on \textit{mini}Imagenet with different numbers of EGNN layers and different feature types}
\vspace{-0.2cm}
\label{table:ablation study}
\end{center}
\end{table}

\subsection{Ablation studies} \label{subsec: ablation_studies}
The proposed edge-labeling GNN has a deep architecture that consists of several node and edge-update layers. Therefore, as the model gets deeper with more layers, the interactions between task samples should be propagated more intensively, which may leads to performance improvements. 
To support this statement, we compared the few-shot learning performances with different numbers of EGNN layers, and the results are presented in Table \ref{table:ablation study}. As the number of EGNN layers increases, the performance gets better. There exists a big jump on few-shot accuracy when the number of layers changes from 1 to 2 (67.99\% $\rightarrow$ 73.19\%), and a little additional gain with three layers (76.37 \%).

Another key ingredient of the proposed EGNN is to use separate exploitation of intra-cluster similarity and inter-cluster dissimilarity in node/edge updates. To validate the effectiveness of this, we conducted experiment with only intra-cluster aggregation and compared the results with those obtained by using both aggregations. The results are also presented in Table \ref{table:ablation study}. For all EGNN layers, the use of separate inter-cluster aggregation clearly improves the performances.

It should also be noted that compared to the previous node-labeling GNN, the proposed edge-labeling framework is more conducive in solving the few-shot problem under arbitrary meta-test setting, especially when the number of few-shot classes for meta-testing does not match to the one used for meta-training. To validate this statement, we conducted a cross-way experiment with EGNN, and the result is presented in Table \ref{table:cross-way exp}. Here, the model was trained with 5-way 5-shot setting and tested on 10-way 5-shot setting, and vice versa. 
Interestingly, both cross-way results are similar to those obtained with the matched-way settings. Therefore, we can observe that the EGNN can be successfully extended to modified few-shot setting without re-training of the model, while the previous node-labeling GNN \cite{garcia2017few} is not even applicable to cross-way setting, since the size of the model and parameters are dependent on the number of ways. 
\begin{table}[]
\begin{center}
\begin{tabular}{l|cc|c}
\hline
Model & Train way & Test way & Accuracy \\ \hline
Prototypical \cite{snell2017prototypical} & 5& 5& 65.77\\
{\bf Prototypical}  & 5& 10& 51.93\\
Prototypical  & 10& 10& 49.29\\
{\bf Prototypical}  & 10& 5& 66.93\\

GNN \cite{garcia2017few}   & 5        & 5       & 66.41    \\
{\bf GNN}    & 5       & 10       & N/A    \\
GNN   & 10        & 10       & 51.75    \\
{\bf GNN}   & 10         & 5       & N/A      \\
\hline
EGNN   & 5         & 5       & 76.37    \\
{\bf EGNN} & 5   & 10       & 56.35    \\
EGNN   & 10         & 10       & 57.61     \\
{\bf EGNN} & 10   & 5       & 76.27    \\ \hline
\end{tabular}
\caption{Cross-way few-shot learning results on miniImagenet 5-shot setting.}
\label{table:cross-way exp}
\end{center}
\end{table}

\begin{figure}[t]
\centering
\includegraphics[width=0.47\textwidth]{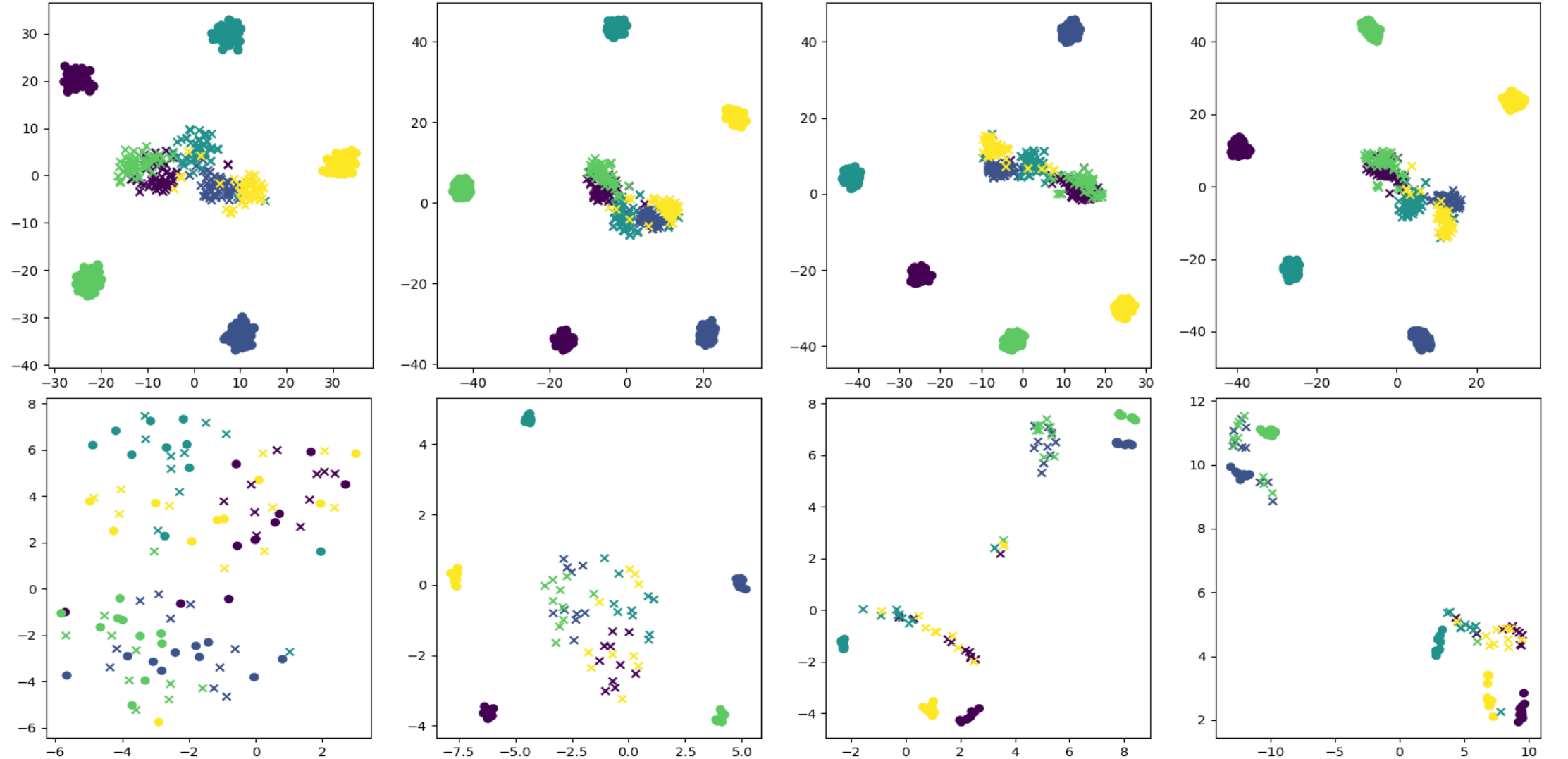}
\caption{t-SNE visualization of node features. From top to bottom: GNN \cite{garcia2017few}, EGNN. From left to right: initial embedding, 1st layer, 2nd layer, 3rd layer. 'x' represents query, 'o' represents support. Different colors mean different labels.}
\vspace{-0.2cm}
\label{fig:tsne}
\end{figure}

Figure \ref{fig:tsne} shows t-SNE \cite{Maaten08} visualizations of node features for the previous node-labeling GNN and EGNN. The GNN tends to show a good clustering among support samples after the first layer-propagation, however, query samples are heavily clustered together, and according to each label, query samples and their support samples never get close together, especially even with more layer-propagations, which means that the last fully-connect layer of GNN actually seems to perform most roles in query classification. In contrast, in our EGNN, as the layer-propagation goes on, both the query and support samples are pulled away if their labels are different, and at the same time, equally labeled query and support samples get close together.

\begin{figure}[t]
\centering
\includegraphics[width=0.47\textwidth]{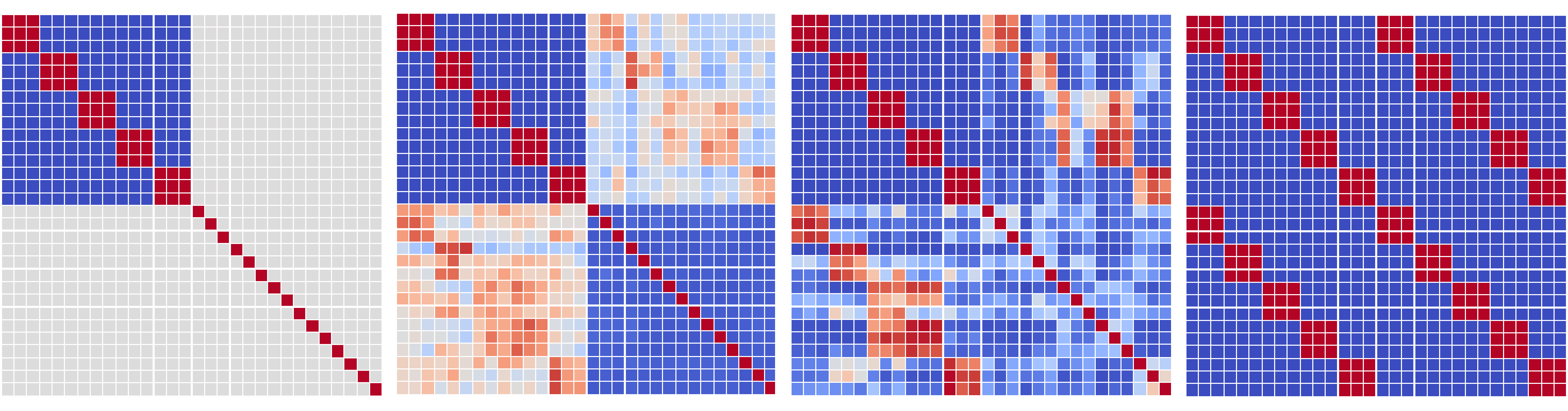}
\caption{Visualization of edge feature propagation. From left to right: initial edge feature, 1st layer, 2nd layer, ground-truth edge labels. Red color denotes higher value ($e_{ij1} = 1$), while blue color denotes lower value ($e_{ij1} = 0$). This illustration shows 5-way 3-shot setting, and 3 queries for each class, total 30 task-samples. The first 15 samples are support set, and latter 15 are query set.} 
\label{fig:heatmap}
\end{figure}

For further analysis, Figure \ref{fig:heatmap} shows how edge features propagate in EGNN. Starting from the initial feature where all query edges are initialized with 0.5, the edge feature gradually evolves to resemble ground-truth edge label, as they are passes through the several EGNN layers. 

\section{Conclusion}
\label{sec:conclusion}
This work addressed the problem of few-shot learning, especially on the few-shot classification task. We proposed the novel EGNN which aims to iteratively update edge-labels for inferring a query association to an existing support clusters. In the process of EGNN, a number of alternative node and edge feature updates were performed using explicit intra-cluster similarity and inter-cluster dissimilarity through the graph layers having different parameter sets, and the edge-label prediction was obtained from the final edge feature. The edge-labeling loss was used to update the parameters of the EGNN with episodic training. Experimental results showed that the proposed EGNN outperformed other few-shot learning algorithms on both of the supervised and semi-supervised few-shot image classification tasks. The proposed framework is applicable to a broad variety of other meta-clustering tasks. 
For future work, we can consider another training loss which is related to the valid graph clustering such as the cycle loss \cite{kim2011higher}. Another promising direction is graph sparsification, e.g. constructing $K$-nearest neighbor graphs \cite{qi20173d}, that will make our algorithm more scalable to larger number of shots.


\section*{Acknowledgement}
This work was supported by the National Research Foundation of Korea (NRF) grant funded by the Korea government (MSIT)(No. NRF-2017R1A2B2006165) and  Institute for Information \& communications Technology Promotion(IITP) grant funded by the Korea government(MSIT) (No.2016-0-00563, Research on Adaptive Machine Learning Technology Development for Intelligent Autonomous Digital Companion). 
Also, we thank the Kakao Brain Cloud team for supporting to efficiently use GPU clusters for large-scale experiments.

{
\small
\bibliographystyle{ieee_fullname}


\begin{thebibliography}{50}
\providecommand{\natexlab}[1]{#1}
\providecommand{\url}[1]{\texttt{#1}}
\expandafter\ifx\csname urlstyle\endcsname\relax
  \providecommand{\doi}[1]{doi: #1}\else
  \providecommand{\doi}{doi: \begingroup \urlstyle{rm}\Url}\fi

\bibitem[Lemke et~al.(2015)Lemke, Budka, and Gabrys]{Lemke15}
Christiane Lemke, Marcin Budka, and Bogdan Gabrys.
\newblock Metalearning: a survey of trends and technologies.
\newblock \emph{Artificial Intelligence Review}, 44\penalty0 (1), 2015.

\bibitem[Vinyals et~al.(2016)Vinyals, Blundell, Lillicrap, Wierstra,
  et~al.]{vinyals2016matching}
Oriol Vinyals, Charles Blundell, Tim Lillicrap, Daan Wierstra, et~al.
\newblock Matching networks for one shot learning.
\newblock In \emph{NIPS}, pages 3630--3638, 2016.

\bibitem[Snell et~al.(2017)Snell, Swersky, and Zemel]{snell2017prototypical}
Jake Snell, Kevin Swersky, and Richard Zemel.
\newblock Prototypical networks for few-shot learning.
\newblock In \emph{NIPS}, pages 4077--4087, 2017.

\bibitem[Finn et~al.(2017)Finn, Abbeel, and Levine]{finn2017model}
Chelsea Finn, Pieter Abbeel, and Sergey Levine.
\newblock Model-agnostic meta-learning for fast adaptation of deep networks.
\newblock In \emph{ICML}, 2017.

\bibitem[Yang et~al.(2018)Yang, Zhang, Xiang, Torr, and
  Hospedales]{yang2018learning}
Flood Sung~Yongxin Yang, Li~Zhang, Tao Xiang, Philip~HS Torr, and Timothy~M
  Hospedales.
\newblock Learning to compare: Relation network for few-shot learning.
\newblock In \emph{CVPR}, 2018.

\bibitem[Garcia and Bruna(2018)]{garcia2017few}
Victor Garcia and Joan Bruna.
\newblock Few-shot learning with graph neural networks.
\newblock In \emph{ICLR}, 2018.

\bibitem[Ren et~al.(2018)Ren, Triantafillou, Ravi, Snell, Swersky, Tenenbaum,
  Larochelle, and Zemel]{ren2018meta}
Mengye Ren, Eleni Triantafillou, Sachin Ravi, Jake Snell, Kevin Swersky,
  Joshua~B Tenenbaum, Hugo Larochelle, and Richard~S Zemel.
\newblock Meta-learning for semi-supervised few-shot classification.
\newblock In \emph{ICLR}, 2018.

\bibitem[Ravi and Larochelle(2017)]{ravi2016optimization}
Sachin Ravi and Hugo Larochelle.
\newblock Optimization as a model for few-shot learning.
\newblock In \emph{ICLR}, 2017.

\bibitem[Santoro et~al.(2016)Santoro, Bartunov, Botvinick, Wierstra, and
  Lillicrap]{Santoro2016mann}
Adam Santoro, Sergey Bartunov, Matthew Botvinick, Daan Wierstra, and Timothy
  Lillicrap.
\newblock Meta-learning with memory-augmented neural networks.
\newblock In \emph{ICML}, pages 1842--1850, 2016.

\bibitem[Mishra et~al.(2018)Mishra, Rohaninejad, Chen, and
  Abbeel]{MishraICLR18}
Nikhil Mishra, Mostafa Rohaninejad, Xi~Chen, and Pieter Abbeel.
\newblock A simple neural attentive meta-learner.
\newblock In \emph{ICLR}, 2018.

\bibitem[Oreshkin et~al.(2018)Oreshkin, Rodriguez, and Lacoste]{Borisnips18}
Boris~N. Oreshkin, Pau Rodriguez, and Alexandre Lacoste.
\newblock Tadam: Task dependent adaptive metric for improved few-shot learning.
\newblock In \emph{NIPS}, 2018.

\bibitem[Liu et~al.(2019)Liu, Lee, Park, Kim, and Yang]{Yanbin18}
Yanbin Liu, Juho Lee, Minseop Park, Saehoon Kim, and Yi~Yang.
\newblock Transductive propagation network for few-shot learning.
\newblock In \emph{ICLR}, 2019.

\bibitem[Wang et~al.(2018)Wang, Girshick, Hebert, and Hariharan]{Wang18}
Yu{-}Xiong Wang, Ross~B. Girshick, Martial Hebert, and Bharath Hariharan.
\newblock Low-shot learning from imaginary data.
\newblock In \emph{CVPR}, 2018.

\bibitem[Lake et~al.(2015)Lake, Salakhutdinov, and tenenbaum]{Lake15}
Brenden~M Lake, Ruslan Salakhutdinov, and Joshua~B tenenbaum.
\newblock Human-level concept learning through probabilistic program induction.
\newblock \emph{Science}, 350\penalty0 (6266):\penalty0 1332--1338, 2015.

\bibitem[Kim et~al.(2018)Kim, Yoon, Dia, Kim, Bengio, and Ahn]{kim2018bayesian}
Taesup Kim, Jaesik Yoon, Ousmane Dia, Sungwoong Kim, Yoshua Bengio, and Sungjin
  Ahn.
\newblock Bayesian model-agnostic meta-learning.
\newblock In \emph{NIPS}, 2018.

\bibitem[Andrychowicz et~al.(2016)Andrychowicz, Denil, Colmenarejo, Hoffman,
  Pfau, Schaul, and de~Freitas]{AndrychowiczDGH16}
Marcin Andrychowicz, Misha Denil, Sergio~Gomez Colmenarejo, Matthew~W. Hoffman,
  David Pfau, Tom Schaul, and Nando de~Freitas.
\newblock Learning to learn by gradient descent by gradient descent.
\newblock In \emph{NIPS}, 2016.

\bibitem[Bello et~al.(2017)Bello, Zoph, Vasudevan, and Le]{Bello17}
Irwan Bello, Barret Zoph, Vijay Vasudevan, and Quoc~V. Le.
\newblock Neural optimizer search with reinforcement learning.
\newblock In \emph{ICML}, 2017.

\bibitem[Wichrowska et~al.(2017)Wichrowska, Maheswaranathan, Hoffman,
  Colmenarejo, Denil, de~Freitas, and Sohl-Dickstein]{Wichrowska17}
Olga Wichrowska, Niru Maheswaranathan, Matthew~W. Hoffman, Sergio~Gomez
  Colmenarejo, Misha Denil, Nando de~Freitas, and Jascha Sohl-Dickstein.
\newblock Learned optimizers that scale and generalize.
\newblock In \emph{ICML}, 2017.

\bibitem[Al{-}Shedivat et~al.(2018)Al{-}Shedivat, Bansal, Burda, Sutskever,
  Mordatch, and Abbeel]{MaruanICLR18}
Maruan Al{-}Shedivat, Trapit Bansal, Yuri Burda, Ilya Sutskever, Igor Mordatch,
  and Pieter Abbeel.
\newblock Continuous adaptation via meta-learning in nonstationary and
  competitive environments.
\newblock In \emph{ICLR}, 2018.

\bibitem[Houthooft et~al.(2018)Houthooft, Chen, Isola, Stadie, Wolski, Ho, and
  Abbeel]{Houthoofnips18}
Rein Houthooft, Richard~Y. Chen, Phillip Isola, Bradly~C. Stadie, Filip Wolski,
  Jonathan Ho, and Pieter Abbeel.
\newblock Evolved policy gradients.
\newblock In \emph{NIPS}, 2018.

\bibitem[Clavera et~al.(2018)Clavera, Nagabandi, Fearing, Abbeel, Levine, and
  Finn]{Clavera18}
Ignasi Clavera, Anusha Nagabandi, Ronald~S. Fearing, Pieter Abbeel, Sergey
  Levine, and Chelsea Finn.
\newblock Learning to adapt: Meta-learning for model-based control.
\newblock \emph{CoRR}, abs/1803.11347, 2018.
\newblock URL \url{http://arxiv.org/abs/1803.11347}.

\bibitem[Vuorio et~al.(2018)Vuorio, Cho, Kim, and Kim]{Vuorio18}
Risto Vuorio, Dong-Yeon Cho, Daejoong Kim, and Jiwon Kim.
\newblock Meta continual learning.
\newblock \emph{arXiv}, 2018.
\newblock URL \url{https://arxiv.org/abs/1806.06928}.

\bibitem[Xu and Zhu(2018)]{Ju}
Ju~Xu and Zhanxing Zhu.
\newblock Reinforced continual learning.
\newblock In \emph{NIPS}, 2018.

\bibitem[Battaglia et~al.(2018)]{Battaglia18}
Peter~W. Battaglia et~al.
\newblock Relational inductive biases, deep learning, and graph networks.
\newblock \emph{arXiv}, 2018.
\newblock URL \url{https://arxiv.org/abs/1806.01261}.

\bibitem[Bronstein et~al.(2017)Bronstein, Bruna, LeCun, Szlam, and
  Vandergheynst]{Bronstein}
Michael~M. Bronstein, Joan Bruna, Yann LeCun, Arthur Szlam, and Pierre
  Vandergheynst.
\newblock Geometric deep learning: going beyond euclidean data.
\newblock \emph{IEEE Signal Processing Magazine}, 34\penalty0 (4):\penalty0
  18--42, 2017.

\bibitem[Xu et~al.(2018)Xu, Hu, Leskovec, and Jegelka]{xu2018powerful}
Keyulu Xu, Weihua Hu, Jure Leskovec, and Stefanie Jegelka.
\newblock How powerful are graph neural networks?
\newblock \emph{arXiv preprint arXiv:1810.00826}, 2018.

\bibitem[Gilmer et~al.(2017)Gilmer, Schoenholz, Riley, Vinyals, and
  Dahl]{GilmerSRVD17}
Justin Gilmer, Samuel~S. Schoenholz, Patrick~F. Riley, Oriol Vinyals, and
  George~E. Dahl.
\newblock Neural message passing for quantum chemistry.
\newblock \emph{CoRR}, abs/1704.01212, 2017.
\newblock URL \url{http://arxiv.org/abs/1704.01212}.

\bibitem[Gori et~al.(2005)Gori, Monfardini, and Scarselli]{Gori05}
M.~Gori, G.~Monfardini, and F.~Scarselli.
\newblock A new model for learning in graph domains.
\newblock In \emph{IJCNN}, 2005.

\bibitem[Scarselli et~al.(2008)Scarselli, Gori, Tsoi, Hagenbuchner, and
  Monfardini]{Scarselli09}
Franco Scarselli, Marco Gori, Ah~Chung Tsoi, Markus Hagenbuchner, and Gabriele
  Monfardini.
\newblock The graph neural network model.
\newblock \emph{IEEE Transactions on Neural Networks}, 20\penalty0
  (1):\penalty0 61--80, 2008.

\bibitem[Kipf and Welling(2017)]{Kipf17}
Thomas~N. Kipf and Max Welling.
\newblock Semi-supervised classification with graph convolutional networks.
\newblock In \emph{ICLR}, 2017.

\bibitem[Li et~al.(2016)Li, Tarlow, Brockschmidt, and Zemel]{Li16}
Yujia Li, Daniel Tarlow, Marc Brockschmidt, and Richard Zemel.
\newblock Gated graph sequence neural networks.
\newblock In \emph{ICLR}, 2016.

\bibitem[Hamilton et~al.(2017)Hamilton, Ying, and Leskovec]{Hamilton17}
William~L. Hamilton, Rex Ying, and Jure Leskovec.
\newblock Inductive representation learning on large graphs.
\newblock In \emph{NIPS}, 2017.

\bibitem[Velickovic et~al.(2018)Velickovic, Cucurull, Casanova, Romero, Lio,
  and Bengio]{Velickovic18}
Petar Velickovic, Guillem Cucurull, Arantxa Casanova, Adriana Romero, Pietro
  Lio, and Yoshua Bengio.
\newblock Graph attention networks.
\newblock In \emph{ICLR}, 2018.

\bibitem[Defferrard et~al.(2016)Defferrard, Bresson, and
  Vandergheynst]{Defferrard16}
Michael Defferrard, Xavier Bresson, and Pierre Vandergheynst.
\newblock Convolutional neural networks on graphs with fast localized spectral
  filtering.
\newblock In \emph{NIPS}, 2016.

\bibitem[Kim et~al.(2011)Kim, Nowozin, Kohli, and Yoo]{kim2011higher}
Sungwoong Kim, Sebastian Nowozin, Pushmeet Kohli, and Chang~D Yoo.
\newblock Higher-order correlation clustering for image segmentation.
\newblock In \emph{NIPS}, pages 1530--1538, 2011.

\bibitem[Gong and Cheng(2018)]{gong2018adaptive}
Liyu Gong and Qiang Cheng.
\newblock Adaptive edge features guided graph attention networks.
\newblock \emph{arXiv preprint arXiv:1809.02709}, 2018.

\bibitem[Kipf et~al.(2018)Kipf, Fetaya, Wang, Welling, and
  Zemel]{kipf2018neural}
Thomas Kipf, Ethan Fetaya, Kuan-Chieh Wang, Max Welling, and Richard Zemel.
\newblock Neural relational inference for interacting systems.
\newblock \emph{arXiv preprint arXiv:1802.04687}, 2018.

\bibitem[Bruna et~al.(2013)Bruna, Zaremba, Szlam, and LeCun]{Bruna13}
Joan Bruna, Wojciech Zaremba, Arthur Szlam, and Yann LeCun.
\newblock Spectral networks and locally connected networks on graphs.
\newblock \emph{CoRR}, abs/1312.6203, 2013.

\bibitem[Henaff et~al.(2015)Henaff, Bruna, and LeCun]{Henaff15}
Mikael Henaff, Joan Bruna, and Yann LeCun.
\newblock Deep convolutional networks on graph-structured data.
\newblock \emph{CoRR}, abs/1506.05163, 2015.

\bibitem[Bansal et~al.(2004)Bansal, Blum, and Chawla]{Bansal04}
N.~Bansal, A.~Blum, and S.~Chawla.
\newblock Correlation clustering.
\newblock \emph{Machine Learning}, 56:\penalty0 89--113, 2004.

\bibitem[Finley and Joachims(2005)]{Finley05}
T.~Finley and T.~Joachims.
\newblock Supervised clustering with support vector machines.
\newblock In \emph{ICML}, 2005.

\bibitem[Taskar(2004)]{Taskar04}
B.~Taskar.
\newblock Learning structured prediction models: a large margin approach.
\newblock \emph{Ph.D. thesis, Stanford University}, 2004.

\bibitem[Johnson(2016)]{johnson2016learning}
Daniel~D Johnson.
\newblock Learning graphical state transitions.
\newblock In \emph{ICLR}, 2016.

\bibitem[Koch et~al.(2015)Koch, Zemel, and Salakhutdinov]{Koch15}
Gregory Koch, Richard Zemel, and Ruslan Salakhutdinov.
\newblock Siamese neural networks for one-shot image recognition.
\newblock 2015.

\bibitem[Zhou et~al.(2018)Zhou, Wu, and Li]{DEML}
Fengwei Zhou, Bin Wu, and Zhenguo Li.
\newblock Deep meta-learning: Learning to learn in the concept space.
\newblock \emph{CoRR}, abs/1802.03596, 2018.

\bibitem[Nichol et~al.(2018)Nichol, Achiam, and Schulman]{reptile}
Alex Nichol, Joshua Achiam, and John Schulman.
\newblock On first-order meta-learning algorithms.
\newblock \emph{CoRR}, abs/1803.02999, 2018.

\bibitem[Russakovsky et~al.(2015)Russakovsky, Deng, Su, Krause, Satheesh, Ma,
  Huang, Karpathy, Khosla, Bernstein, et~al.]{russakovsky2015imagenet}
Olga Russakovsky, Jia Deng, Hao Su, Jonathan Krause, Sanjeev Satheesh, Sean Ma,
  Zhiheng Huang, Andrej Karpathy, Aditya Khosla, Michael Bernstein, et~al.
\newblock Imagenet large scale visual recognition challenge.
\newblock \emph{International Journal of Computer Vision}, 115\penalty0
  (3):\penalty0 211--252, 2015.

\bibitem[Paszke et~al.(2017)Paszke, Gross, Chintala, Chanan, Yang, DeVito, Lin,
  Desmaison, Antiga, and Lerer]{paszke2017automatic}
Adam Paszke, Sam Gross, Soumith Chintala, Gregory Chanan, Edward Yang, Zachary
  DeVito, Zeming Lin, Alban Desmaison, Luca Antiga, and Adam Lerer.
\newblock Automatic differentiation in pytorch.
\newblock In \emph{NIPS-W}, 2017.

\bibitem[van~der Maaten and Hinton(2008)]{Maaten08}
L.~van~der Maaten and G.~Hinton.
\newblock Visualizing data using t-sne.
\newblock \emph{JMLR}, 9:\penalty0 2579--2605, 2008.

\bibitem[Qi et~al.(2017)Qi, Liao, Jia, Fidler, and Urtasun]{qi20173d}
Xiaojuan Qi, Renjie Liao, Jiaya Jia, Sanja Fidler, and Raquel Urtasun.
\newblock 3d graph neural networks for rgbd semantic segmentation.
\newblock In \emph{Proceedings of the IEEE International Conference on Computer
  Vision}, pages 5199--5208, 2017.

\end{thebibliography}
}

\end{document}